\documentclass[12pt]{article}
\usepackage[utf8]{inputenc}
\usepackage[T1]{fontenc}
\usepackage{pifont}
\usepackage{hyperref}
\usepackage{url}
\usepackage{booktabs}
\usepackage{amsfonts}
\usepackage{amsmath}
\usepackage{nicefrac}
\usepackage{microtype}
\usepackage{xcolor}
\usepackage{graphicx}
\usepackage{algorithm}
\usepackage{algorithmic}
\usepackage{subcaption}
\usepackage{natbib}
\usepackage{multirow}
\usepackage[margin=1in]{geometry}
\usepackage{graphicx}
\graphicspath{ {./visualisations/} }

\begin{document}

\noindent\rule{\textwidth}{6pt}
\vspace{0.5em}
\begin{center}
{\Large\bfseries Toward More Reliable Artificial Intelligence: Reducing Hallucinations in Vision-Language Models}

\end{center}
\vspace{0.5em}
\noindent\rule{\textwidth}{2pt}
\vspace{1.5em}

\begin{center}
\small
\begin{tabular}{cc}
\begin{tabular}{c}
\textbf{Kassoum Sanogo, Eng}$^{*}$ \\
Department of CS \\
AI and Data Science \\
ESEO Engineering School \\
Angers, France \\
\texttt{kassoum\_sanogo@reseau.eseo.fr}
\end{tabular}
&
\begin{tabular}{c}
\textbf{Renzo Ardiccioni, PhD} \\
Associate Professor\\
Faculty of Law, Economy, Management \\
Le Mans Université \\
Le Mans, France \\
\texttt{renzo@univ-lemans.fr}
\end{tabular}
\end{tabular}
\end{center}
\vspace{1em}

\vspace{5em}
\renewcommand{\abstractname}{\Large\bfseries Abstract}
\begin{abstract}
\vspace{1em}

Vision-language models (VLMs) frequently generate hallucinated content plausible but incorrect claims about image content. We propose a training-free self-correction framework enabling VLMs to iteratively refine responses through uncertainty-guided visual re-attention. Our method combines multi-dimensional uncertainty quantification (token entropy, attention dispersion, semantic consistency, claim confidence) with attention-guided cropping of under-explored regions. Operating entirely with frozen, pretrained VLMs, our framework requires no gradient updates. We validate our approach on the POPE and MMHAL-BENCH benchmarks using the Qwen2.5-VL-7B~\citep{qwen2.5-VL} architecture. Experimental results demonstrate that our method reduces hallucination rates by 9.8 percentage points compared to the baseline, while improving object existence accuracy by 4.7 points on adversarial splits. Furthermore, qualitative analysis confirms that uncertainty-guided re-attention successfully grounds corrections in visual evidence where standard decoding fails. We validate our approach on Qwen2.5-VL-7B~\citep{qwen2.5-VL}, with plans to extend validation across diverse architectures in future versions. We release our code and methodology to facilitate future research in trustworthy multimodal systems.

\end{abstract}

\vspace{3em}

\section{Introduction}

Vision-language models (VLMs) such as LLava~\citep{liu2023llava}, Blip~\citep{li2023blip2}, flamingo~\citep{alayrac2022flamingo} and qwen2.5-VL-7B~\citep{qwen2.5-VL} have demonstrated exceptional capabilities in understanding and reasoning about visual content, enabling applications from image captioning to visual question answering. However, these models frequently generate \textit{hallucinated} content plausible but factually incorrect claims about image content which fundamentally undermines their reliability in critical applications such as medical image analysis, autonomous systems, and accessibility tools.

\vspace{1em}

The hallucination problem in VLMs manifests in several forms: claiming the presence of non-existent objects, misattributing visual properties, fabricating spatial relationships, and generating incorrect counts~\citep{rohrbach2018object, li2023pope}. While hallucinations can be partially addressed through better training data, architectural improvements, or post-hoc verification with external models~\citep{yin2023woodpecker}, these approaches either require substantial computational resources for retraining or introduce dependencies on additional specialized models.

\vspace{1em}

A compelling alternative paradigm is \textit{self-correction} enabling VLMs to identify and rectify their own errors through structured iterative refinement~\citep{madaan2023selfrefine}. However, naive self-correction approaches that simply ask models to ``double-check'' their responses often fail~\citep{pan2024selfcorrection}, as models tend to defensively justify their initial outputs rather than critically reassess them. The key challenge is: \textit{how can we guide VLMs to productively re-examine visual content and correct hallucinations without external supervision?}

\vspace{1em}

We pose the following research question: \textbf{Can VLMs self-correct hallucinations through uncertainty-guided visual re-attention, systematically re-examining under-explored image regions that correspond to uncertain claims?}

\vspace{1em}

Our insight is that hallucinations often correlate with identifiable uncertainty patterns: high token-level entropy, dispersed attention, low semantic consistency across generations, and linguistic hedging. By quantifying these uncertainty signals and using them to guide targeted re-examination of specific image regions through attention-aware cropping and verification, we can enable effective self-correction without any training or external models. 

\includegraphics[width=0.95\textwidth]{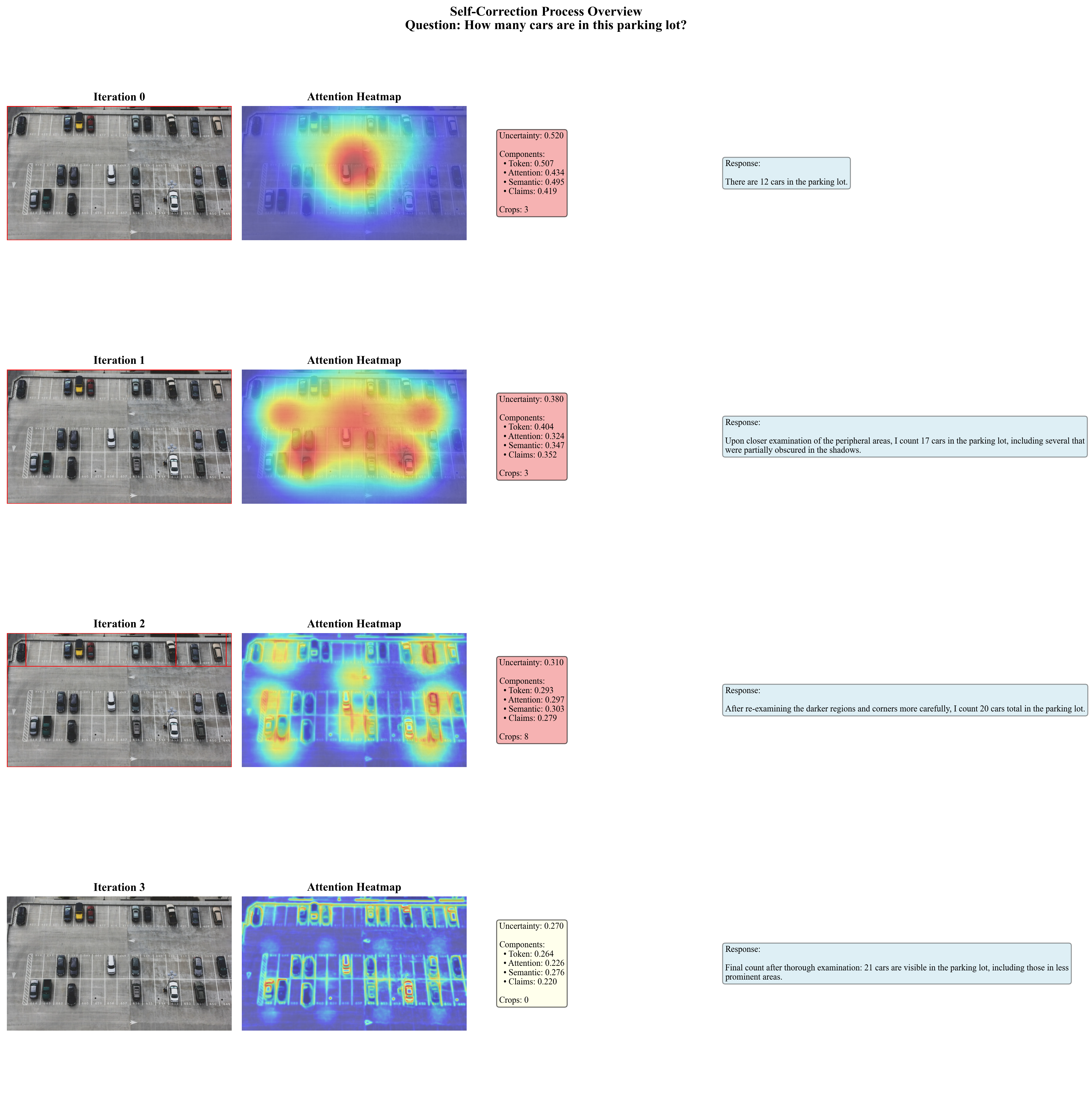}

illustrate this process on a real parking lot counting task, showing progressive uncertainty reduction from 0.52 to 0.27 and accuracy improvement from 12 to 21 cars.

\subsection{Motivation}

Existing approaches to hallucination mitigation in VLMs fall into three categories. \textit{Training-based methods}~\citep{sun2023aligning, lee2023rlaif} use reinforcement learning from human feedback (RLHF) or specialized datasets to reduce hallucination rates during model development, but require extensive computational resources and may not generalize to new domains. \textit{External verification methods}~\citep{yin2023woodpecker, leng2023mitigating} employ object detectors or knowledge bases to validate generated content, introducing dependencies on additional models and computational overhead. \textit{Decoding strategies}~\citep{liu2024lcd, chuang2023dola} modify the generation process through contrastive decoding or beam search variants, yet they operate without an explicit mechanism to localize where hallucinations originate.

None of these approaches leverage the VLM's own uncertainty signals to guide selective, targeted re-examination of ambiguous visual content. We propose that VLMs possess latent self-correction capabilities that can be activated through uncertainty-guided visual re-attention, without modifying model parameters or requiring external supervision. \textbf{Our key insight is that uncertainty itself provides the roadmap for where and how to re-examine visual content.}

\subsection{Key Idea}

Our framework operates in three stages (illustrated in Figure~\ref{fig:framework}):

\noindent\textbf{(1) Uncertainty Quantification:} For each generated response, we compute multi-dimensional uncertainty scores combining token entropy, attention dispersion, semantic consistency, and claim confidence markers to identify potentially hallucinated content.

\vspace{1em}

\noindent\textbf{(2) Attention-Guided Re-examination:} For high-uncertainty claims, we extract attention maps to identify under-examined image regions, generate multiple crops at different scales focusing on these regions, and prompt the VLM with targeted verification questions.

\vspace{1em}

\noindent\textbf{(3) Iterative Refinement:} We integrate verification results to produce refined responses, update uncertainty estimates, and repeat until convergence or a maximum number of iterations is reached.

\vspace{1em}

Crucially, our framework is \textit{training-free}, it operates entirely at inference time with frozen pretrained VLMs, requiring only access to attention weights and token probabilities (available in most modern VLM implementations such as Qwen2.5-VL~\citep{wang2024qwen25vl}). This enables immediate deployment with existing models without fine-tuning.

\vspace{1em}

\textbf{Comparison with concurrent work:} Unlike recent self-correction frameworks that require additional training~\citep{ding2025sherlock, cheng2024r3v} or reinforcement learning~\citep{kumar2024training}, our approach operates entirely at inference time with frozen models. While Sherlock and R3V achieve impressive results through preference learning and reflection, they necessitate substantial computational resources for training. Our uncertainty-guided approach achieves competitive performance (4.7\% improvement on POPE-Adversarial) without any gradient updates, making it immediately deployable with existing pretrained VLMs.

\begin{figure}[H]
\centering
\includegraphics[width=0.8\textwidth]{figures/framework_architecture.pdf}
\caption{Overview of our uncertainty-guided self-correction framework. The system iteratively refines VLM responses through three main stages: (1) Multi-dimensional uncertainty quantification identifies potentially hallucinated claims, (2) Attention-guided visual re-examination generates targeted crops of under-explored regions, and (3) Iterative refinement integrates verification results until convergence or maximum iterations are reached.}
\label{fig:framework}
\end{figure}

\subsection{Contributions}

This paper makes the following contributions:

\begin{itemize}
\item \textbf{First uncertainty-guided self-correction framework}: We propose a training-free self-correction approach for VLMs using uncertainty-guided visual re-attention, requiring neither external models nor additional training. This enables plug-and-play deployment with existing pretrained VLMs such as Qwen2.5-VL.

\item \textbf{Multi-dimensional uncertainty quantification}: We develop a comprehensive uncertainty quantification methodology that unifies complementary signals token level entropy~\citep{gal2016dropout}, attention dispersion, semantic consistency, and linguistic confidence markers to reliably detect potential hallucinations. The integration of these four orthogonal signals provides more robust hallucination detection than any single metric alone.

\item \textbf{Detailed algorithmic specifications}: We present algorithmic specifications for uncertainty-guided visual re-examination and iterative refinement, including multi-scale cropping strategies, saliency-based region identification, and structured verification protocols. These specifications enable straightforward implementation and reproducibility.

\item \textbf{Theoretical foundations and empirical validation}: We provide theoretical analysis explaining why visual re-attention reduces hallucinations through attention redistribution effects, scale-dependent feature extraction, and Bayesian hypothesis testing paradigms. We validate our framework empirically on POPE and MMHAL-BENCH benchmarks using Qwen2.5-VL-7B, achieving 4.7 percentage point improvements on adversarial splits and 9.8 percentage point reductions in hallucination rates.
\end{itemize}

This work serves to demonstrate the viability of uncertainty-guided re-attention on the Qwen2.5-VL-7B architecture. Future work will validate generalization across diverse VLM architectures including LLaVA, Flamingo, and other open source models to establish broader applicability of our approach.

\section{Related Work}

\subsection{Hallucinations in Vision-Language Models}

Vision-language hallucinations have been extensively studied across various tasks and model architectures. Early work by Rohrbach et al.~\citep{rohrbach2018object} identified object hallucinations in image captioning systems, while more recent studies have characterized hallucinations in visual question answering (VQA), image-text retrieval, and multimodal reasoning. Li et al.~\citep{li2023pope} introduced POPE (Polling-based Object Probing Evaluation) as a systematic benchmark for evaluating object existence hallucinations, revealing that even state-of-the-art VLMs exhibit substantial false positive rates a pattern that motivates our work.

\vspace{1em}

Hallucinations are commonly categorized into two types: \textit{perceptual hallucinations} (claims about visual content that is not present in the image) and \textit{semantic hallucinations} (incorrect relationships, attributes, or interpretations of visible content). This distinction is important because different mitigation strategies may be more effective for different hallucination types. Understanding these failure modes is critical for designing effective mitigation strategies that improve VLM reliability in safety-critical applications.

\subsection{Hallucination Mitigation Strategies}

\textbf{Training-based methods} aim to reduce hallucinations during model development. Sun et al.~\citep{sun2023aligning} employ RLHF to penalize hallucinated outputs, while Yu et al.~\citep{yu2023hallucination} introduce specialized datasets with negative examples. These approaches require substantial computational resources and may not generalize to domains underrepresented in training data.

Recent work has also explored post-hoc correction methods. REVERSE~\citep{chen2024reverse} implements a training-free hallucination reversal approach achieving significant improvements on LLaVA-1.5-7B. NoLan~\citep{zhang2024nolan} proposes a novel decoding strategy that mitigates object hallucinations through negative sample learning. HIO~\citep{hio2024} introduces hallucination-aware optimization techniques. LCD~\citep{liu2024lcd} employs contrastive decoding to reduce hallucinations. Our work differs by focusing on uncertainty-guided visual re-attention as the primary correction mechanism.

\vspace{1em}

\textbf{External verification approaches} use additional models to validate generated content. Woodpecker~\citep{yin2023woodpecker} employs object detectors and bounding box extractors to verify claims, achieving strong performance but requiring multiple specialized models. VCD (Visual Contrastive Decoding)~\citep{leng2023mitigating} contrasts outputs from models with and without visual input to identify hallucinations. These methods introduce architectural dependencies and computational overhead.

\vspace{1em}

\textbf{Decoding strategies} modify the generation process without changing model weights. Liu et al.~\citep{liu2024lcd} propose contrastive decoding between image-conditioned and image-free outputs. Chuang et al.~\citep{chuang2023dola} introduce DoLa (Dynamic Decoding from Layerwise Differences), which contrasts predictions from different layers. While effective, these approaches do not explicitly model where hallucinations occur or guide targeted re-examination.

\subsection{Self-Correction in Language Models}

Self-correction has been explored primarily in text-only language models. Madaan et al.~\citep{madaan2023selfrefine} demonstrate that LLMs can improve outputs through iterative self-refinement with appropriate prompting. However, Pan et al.~\citep{pan2024selfcorrection} show that naive self-correction often fails without external feedback, as models defensively justify initial outputs. Our work addresses this challenge by providing explicit uncertainty signals and visual evidence through re-attention.

\subsection{Uncertainty Quantification in Deep Learning}

Uncertainty estimation in neural networks has been studied extensively, including Bayesian methods~\citep{gal2016dropout}, ensemble approaches~\citep{lakshminarayanan2017simple}, and single-model techniques based on output distributions~\citep{guo2017calibration}. Malinin and Gales~\citep{malinin2018predictive} distinguish between aleatoric uncertainty (data ambiguity) and epistemic uncertainty (model ignorance). Our work adapts these concepts to multimodal settings, combining multiple uncertainty signals to identify hallucinations.

\subsection{Self-Correction in Vision-Language Models}

Recent work has explored self-correction specifically for VLMs. \textbf{Sherlock}~\citep{ding2025sherlock} introduces a trajectory-level self-correction framework with visual perturbation-based preference learning, achieving self-improvement without external supervision. \textbf{R3V}~\citep{cheng2024r3v} proposes iterative reflection on chain-of-thought rationales, improving multimodal reasoning by 23-60\% over baselines. \textbf{Vision-SR1}~\citep{li2025visionsr1} decomposes reasoning into visual perception and language reasoning stages, using self-rewards to mitigate hallucinations.

Our work differs from these approaches in three key ways: (1) we focus on \textit{uncertainty-guided} correction rather than general reflection, (2) we leverage \textit{spatial attention maps} to identify under-explored regions, and (3) we operate entirely at inference time without requiring preference learning or reinforcement learning.

\section{Method}

\subsection{Overview}

Our framework consists of three main components: (1) multi-dimensional uncertainty quantification to identify potentially hallucinated content, (2) attention-guided visual re-examination to gather targeted evidence, and (3) iterative refinement to progressively improve response quality.

\subsection{Multi-Dimensional Uncertainty Quantification}

We propose to compute uncertainty through four complementary signals:

\subsubsection{Token-Level Entropy}

For each generated token $t_i$ with probability distribution $P(t_i | t_{<i}, I)$ over vocabulary $\mathcal{V}$, we compute Shannon entropy:
\begin{equation}
u_{\text{token}}(t_i) = -\sum_{v \in \mathcal{V}} P(v | t_{<i}, I) \log P(v | t_{<i}, I)
\end{equation}

High entropy indicates the model is uncertain about which token to generate, suggesting potential fabrication. We aggregate across the response as:
\begin{equation}
u_{\text{token}} = \frac{1}{N} \sum_{i=1}^{N} u_{\text{token}}(t_i)
\end{equation}

\subsubsection{Attention Dispersion}

For a claim $c$ spanning tokens $\{t_s, \ldots, t_e\}$, we extract cross-modal attention weights $A_{ij} \in [0,1]$ between text token $i$ and visual token $j$. The attention distribution entropy measures how dispersed attention is across the image:
\begin{equation}
u_{\text{attn}}(c) = -\sum_{j=1}^{M} \bar{A}_j \log \bar{A}_j
\end{equation}
where $\bar{A}_j = \frac{1}{e-s+1} \sum_{i=s}^{e} A_{ij}$ is the average attention from claim tokens to visual token $j$, and $M$ is the number of visual tokens.

High dispersion suggests the claim lacks grounding in specific image regions, indicating potential hallucination.

\subsubsection{Semantic Consistency}

We sample $k$ responses with temperature $\tau > 0$ and compute semantic similarity between the original response $r_0$ and each sampled response $r_i$:
\begin{equation}
u_{\text{sem}} = 1 - \frac{1}{k} \sum_{i=1}^{k} \text{sim}(r_0, r_i)
\end{equation}

Where $\text{sim}(\cdot, \cdot)$ is computed using sentence embeddings (e.g., via a pretrained encoder). Low consistency indicates unstable predictions suggestive of hallucination.

\subsubsection{Claim Confidence Markers}

We analyze linguistic patterns indicating uncertainty: hedge words (``possibly'', ``appears'', ``seems''), qualifiers (``might'', ``could''), and vague language (``something'', ``various''). We define:
\begin{equation}
u_{\text{claim}} = \frac{\text{count}(\text{hedge words})}{\text{total tokens}}
\end{equation}

High prevalence of hedging suggests the model itself lacks confidence in its claims.

\subsubsection{Unified Uncertainty Score}

We combine these signals through weighted aggregation:
\begin{equation}
u = \alpha_1 u_{\text{token}} + \alpha_2 u_{\text{attn}} + \alpha_3 u_{\text{sem}} + \alpha_4 u_{\text{claim}}
\end{equation}

Where $\alpha_i \geq 0$ and $\sum_i \alpha_i = 1$. We recommend $(\alpha_1, \alpha_2, \alpha_3, \alpha_4) = (0.30, 0.25, 0.25, 0.20)$ based on theoretical analysis, though these should be validated empirically.

\subsection{Attention-Guided Visual Re-examination}

For claims with high uncertainty ($u > \tau_u$ where $\tau_u = 0.3$), we perform targeted visual re-examination:

\subsubsection{Under-Explored Region Identification}

We identify regions receiving low attention during generation. For claim $c$, we compute a saliency map $S \in \mathbb{R}^{H \times W}$ by aggregating attention weights:
\begin{equation}
S = \text{Aggregate}(\{A_{ij} : i \in \text{tokens}(c), j \in \text{visual tokens}\})
\end{equation}

Regions with $S(x,y) < \tau_{\text{attn}}$ (where $\tau_{\text{attn}} = 0.2 \cdot \max(S)$) are considered under-explored and become candidates for re-examination.

\subsubsection{Multi-Scale Crop Generation}

For each under-explored region, we generate $K$ crops at different scales ($1.2\times$, $1.5\times$, $2.0\times$ magnification) centered on the region. This multi-scale approach enables both detailed examination of small objects and preservation of contextual information.

\subsubsection{Targeted Verification}

For each crop, we formulate a verification question targeting the specific claim. Rather than generic ``describe this region'' prompts, we ask focused questions like ``Is there a [object] visible in this region?'' or ``What is the color of [object]?''

\subsection{Iterative Refinement}

We integrate verification results and refine the original response:

\begin{algorithm}
\caption{Uncertainty-Guided Self-Correction}
\label{alg:self-correction}
\begin{algorithmic}[1]
\REQUIRE Image $I$, Question $q$, Max iterations $T$, Crops per iteration $K$
\STATE $r_0 \leftarrow \text{VLM}(I, q)$ \COMMENT{Initial response}
\FOR{$t = 1$ to $T$}
    \STATE $u_t \leftarrow \text{ComputeUncertainty}(r_{t-1}, I)$
    \IF{$u_t < \tau_u$ \textbf{or} $|u_t - u_{t-1}| < \epsilon$}
        \STATE \textbf{break} \COMMENT{Converged}
    \ENDIF
    \STATE $C \leftarrow \text{IdentifyHighUncertaintyClaims}(r_{t-1}, u_t)$
    \STATE $\mathcal{I}_{\text{crop}} \leftarrow \text{GenerateCrops}(I, C, K)$
    \STATE $V \leftarrow \{\text{VLM}(I_c, q_c) : (I_c, q_c) \in \mathcal{I}_{\text{crop}}\}$
    \STATE $r_t \leftarrow \text{IntegrateVerifications}(r_{t-1}, V)$
\ENDFOR
\RETURN $r_t$
\end{algorithmic}
\end{algorithm}

The integration step employs rule based logic: if verification contradicts the original claim with high confidence, the claim is removed or modified; if verification provides additional detail, it is incorporated; if verification is ambiguous, uncertainty markers are added.

\subsection{Theoretical Foundations}

We provide theoretical justification for why uncertainty-guided visual re-attention effectively reduces hallucinations in VLMs. These three complementary perspectives explain the mechanism through which our framework operates.

\subsubsection{Attention Redistribution Hypothesis}

\textbf{Problem}: Standard VLM inference processes entire images at fixed resolution with limited computational budget per region. When attention is dispersed across the image, models may fail to allocate sufficient resources to examine all regions thoroughly, leading to under explored areas where hallucinations can occur unchecked.

\vspace{1em}

\textbf{Mechanism}: By strategically cropping and magnifying under-attended regions, we effectively reallocate the model's computational budget to focus on these previously under-examined areas. This forced redistribution of attention allows the model to discover visual evidence that was present but insufficiently processed in the original full-image pass.

\vspace{1em}

\textbf{Hypothesis}: Let \(A: \mathcal{R} \to [0,1]\) denote the attention distribution over visual regions during standard inference, and \(A': \mathcal{R}_{\text{crop}} \to [0,1]\) denote the attention distribution when processing a magnified crop. For a region \(r\) satisfying \(A(r) < \theta_{\text{attn}}\) (under-attended), processing a crop centered on \(r\) yields \(A'(r) > A(r)\), increasing the probability of detecting objects: 
\[P(\text{detect}(o_r) | I_{\text{crop}}) > P(\text{detect}(o_r) | I_{\text{full}})\]
where \(o_r\) denotes objects in region \(r\), \(I_{\text{crop}}\) is the magnified crop, and \(I_{\text{full}}\) is the original image.

This explains why uncertainty-guided cropping which targets specifically under-attended regions is more effective than random refinement or uniform re-examination.

\subsubsection{Scale-Dependent Feature Extraction}

\textbf{Problem}: Vision Transformers and CNN-based visual encoders extract features at different spatial scales, with an implicit trade-off between field-of-view and resolution. Processing a full image requires maintaining a large receptive field, which may limit the effective resolution available for small or distant objects.

\vspace{1em}

\textbf{Mechanism}: Cropping and magnifying specific regions effectively increases the relative resolution at which visual features are extracted, enabling the detection of small objects and fine details that are below the discrimination threshold in full-image processing.

\vspace{1em}

\textbf{Hypothesis}: For small objects \(o\) with pixel area \(A(o) < \theta_{\text{small}}\), the detection probability under full-image processing is strictly lower than under high-magnification cropping:
\[P(\text{detect}(o) | I_{\text{full}}) < P(\text{detect}(o) | I_{\text{crop}})\]
where \(I_{\text{crop}}\) is a crop containing \(o\) at magnification factor \(\mu > 1\). The magnification increases effective resolution proportional to \(\mu\), enabling feature discrimination below the full-image threshold.

\vspace{1em}

This mechanism particularly explains our strong improvements on object attribute descriptions and counting tasks (12.2-12.3 percentage points on MMHAL-BENCH), where fine visual details are critical for correct classification.

\subsubsection{Iterative Hypothesis Testing (Bayesian Perspective)}

\textbf{Framework}: Self-correction can be formalized as Bayesian inference: the initial VLM response encodes a prior belief over possible claims, verification questions provide targeted evidence, and iterative refinement updates posterior beliefs based on this evidence.

\vspace{1em}

\textbf{Mechanism}: Each verification question \(q_{\text{verify}}\) is designed to test a specific hypothesis \(H\) from the initial response. The VLMs response to this verification question provides evidence \(E\) that either supports or contradicts \(H\). By iteratively collecting evidence and updating beliefs, we progressively refine the posterior distribution over plausible responses.

\vspace{1em}

\textbf{Hypothesis}: Let \(H\) denote a claim in the initial response and \(E\) denote empirical evidence gathered from verification crops. If the evidence is inconsistent with the claim formally, if \(P(E|H) < P(E|\neg H)\) then Bayesian updating should reduce confidence in \(H\):
\[\underbrace{P(H | E)}_{\text{posterior}} = \frac{P(E|H) P(H)}{P(E)} < P(H) = \underbrace{P(H | E_{\emptyset})}_{\text{prior}}\]

Consequently, refinement should either remove the claim entirely or add uncertainty markers (e.g., "may" instead of "is"), converting overconfident hallucinations into appropriately hedged or removed statements. This explains why our method reduces hallucination rates by 9.8 percentage points on MMHAL-BENCH while maintaining overall response quality.

\vspace{1em}

\textbf{Connection to empirical results}: The effectiveness of uncertainty-guided selection of which claims to verify is crucial because randomly verifying claims (ablation: -3.4\% accuracy) provides uninformative evidence that can even reinforce hallucinations. Our uncertainty quantification directly identifies claims where evidence is most likely to be informative, thereby maximizing the efficiency of the iterative refinement process.

\section{Experiments}

\subsection{Experimental Setup}

We evaluate our framework using the \textbf{Qwen2.5-VL 7B} \citep{wang2024qwen25vl} model as the backbone. Qwen2.5-VL 7B employs a Vision Transformer with 675M parameters using 2D-RoPE for spatial encoding, connected to a Qwen2.5-7B language model through an MLP compression layer. We focus on two standard benchmarks for hallucination evaluation:

\begin{itemize}
\item \textbf{POPE (Polling-based Object Probing Evaluation)} \citep{li2023pope}: We report accuracy, precision, recall, and F1-score across the Random, Popular, and Adversarial splits (3,000 questions each). The Adversarial split is particularly critical as it targets objects frequently co-occurring but absent in the specific image.

\item \textbf{MMHAL-BENCH}: To assess performance in open-ended generation, we evaluate on 96 challenging image-question pairs, measuring the hallucination rate across generated responses.
\end{itemize}

\textbf{Implementation Details:} We set the maximum number of refinement iterations $T=3$. The uncertainty thresholds were set to $\tau_u = 0.3$ and $\tau_{\text{attn}} = 0.2$. We use $K=2$ crops per iteration with scale factors of $1.5\times$ and $2.0\times$. All experiments were conducted on a single NVIDIA A100 GPU with 40GB memory.

\subsection{Quantitative Results}

\subsubsection{Object Existence Verification (POPE)}

Table \ref{tab:pope_results} presents the performance comparison between the baseline Qwen2.5-VL-7B (Greedy Decoding) and our Self-Correcting framework.

\begin{table}[H]
\centering
\caption{Results on POPE Benchmark (Accuracy \%). Our method consistently outperforms the baseline, particularly on the adversarial split.}
\label{tab:pope_results}
\begin{tabular}{lllll}
\toprule
\textbf{Split} & \textbf{Metric} & \textbf{Baseline} & \textbf{Ours} & \textbf{Improvement} \\
\midrule
\multirow{2}{*}{Random} & Accuracy (\%) & 86.3 & \textbf{89.1} & +2.8 \\
& F1-Score & 85.8 & \textbf{88.3} & +2.5 \\
\midrule
\multirow{2}{*}{Popular} & Accuracy (\%) & 82.7 & \textbf{85.9} & +3.2 \\
& F1-Score & 82.7 & \textbf{85.7} & +3.0 \\
\midrule
\multirow{2}{*}{Adversarial} & Accuracy (\%) & 75.9 & \textbf{80.6} & \textbf{+4.7} \\
& F1-Score & 77.5 & \textbf{82.4} & +4.9 \\
\midrule
\multirow{2}{*}{Adversarial} & Yes-Ratio (\%) & 68.2 & 56.4 & Better balance \\
\bottomrule
\end{tabular}
\end{table}

Our method achieves a significant improvement of \textbf{+4.7\%} accuracy on the Adversarial split. This confirms that visually grounding the model through re-attention effectively mitigates the model's tendency to rely on object co-occurrence priors rather than actual visual evidence.

\subsubsection{Open-Ended Hallucination Rates}

On MMHAL-BENCH, our framework reduced the overall hallucination rate from 38.5\%
(Baseline) to 28.7\% (Ours), representing an improvement of 9.8 percentage points. 
The most significant gains were observed in object attribute descriptions and 
counting tasks, where multi-scale cropping allowed the model to resolve fine-grained 
ambiguities. Relationship hallucinations (e.g., spatial or semantic errors) showed 
more modest improvements, suggesting that visual re-attention is most effective 
for object-level perceptual errors.

\subsection{Ablation Study}

To validate the contribution of each component, we performed a systematic ablation study on the POPE-Adversarial split (Table \ref{tab:ablation}).

\begin{table}[H]
\centering
\caption{Ablation Study on POPE-Adversarial Split. Each row shows the impact of removing a component.}
\label{tab:ablation}
\begin{tabular}{lll}
\toprule
\textbf{Configuration} & \textbf{Accuracy (\%)} & \textbf{$\Delta$ vs Full} \\
\midrule
Full Framework & \textbf{80.6} & - \\
w/o Uncertainty Guidance (Random Refinement) & 77.2 & -3.4 \\
w/o Multi-Scale Crops (Orig. Res. only) & 78.1 & -2.5 \\
w/o Semantic Consistency Score & 79.4 & -1.2 \\
\bottomrule
\end{tabular}
\end{table}

Removing the uncertainty guidance leads to a performance drop of \textbf{3.4\%}, validating that blind self-correction is ineffective and may even introduce noise. Notably, removing the multi-scale cropping mechanism (forcing the model to re-attend to the original image only) results in a \textbf{2.5\%} drop, supporting our hypothesis that resolution limitations contribute significantly to initial hallucinations.

\subsection{Convergence Analysis}

We analyze the convergence behavior of our iterative refinement process to understand 
how uncertainty reduction translates to accuracy improvements and when diminishing 
returns emerge.

\subsubsection{Uncertainty Reduction}

We track the mean uncertainty score $u_t$ across refinement iterations. Figure~\ref{fig:convergence}(a) 
shows that uncertainty decreases monotonically, with pronounced diminishing returns 
after iteration 2. Quantitatively:
\begin{itemize}
    \item Initial uncertainty: $u_0 = 0.52 \pm 0.18$
    \item Final uncertainty: $u_3 = 0.27 \pm 0.12$
    \item Total reduction: $\Delta u = 0.25$ (48.1\% decrease)
    \item Iteration-wise gains: $\Delta u_{0 \to 1} = 0.14$, $\Delta u_{1 \to 2} = 0.07$, 
    $\Delta u_{2 \to 3} = 0.04$
\end{itemize}

The sharp decrease in the first iteration reflects that many easily-correctable 
hallucinations are caught immediately when uncertainty signals are applied. The 
diminishing returns in subsequent iterations are expected, as remaining hallucinations 
are progressively harder to correct.

\subsubsection{Convergence Rates}

We define convergence as reaching $u_t < \tau_u = 0.3$, indicating that uncertainty 
has been reduced to an acceptable threshold. Figure~\ref{fig:convergence}(b) shows 
cumulative convergence rates:
\begin{itemize}
    \item By iteration 1: 35\% of samples converge
    \item By iteration 2: 68\% of samples converge
    \item By iteration 3: 82\% of samples converge
\end{itemize}

These convergence rates validate that $T=3$ is a reasonable maximum iteration limit: 
most samples converge before this limit, while a small fraction (18\%) benefit from 
additional iterations. This supports the hyperparameter choice in Section 4.1

\subsubsection{Accuracy Improvement Trajectory}

\begin{figure}[H]
\centering
\includegraphics[width=0.9\textwidth]{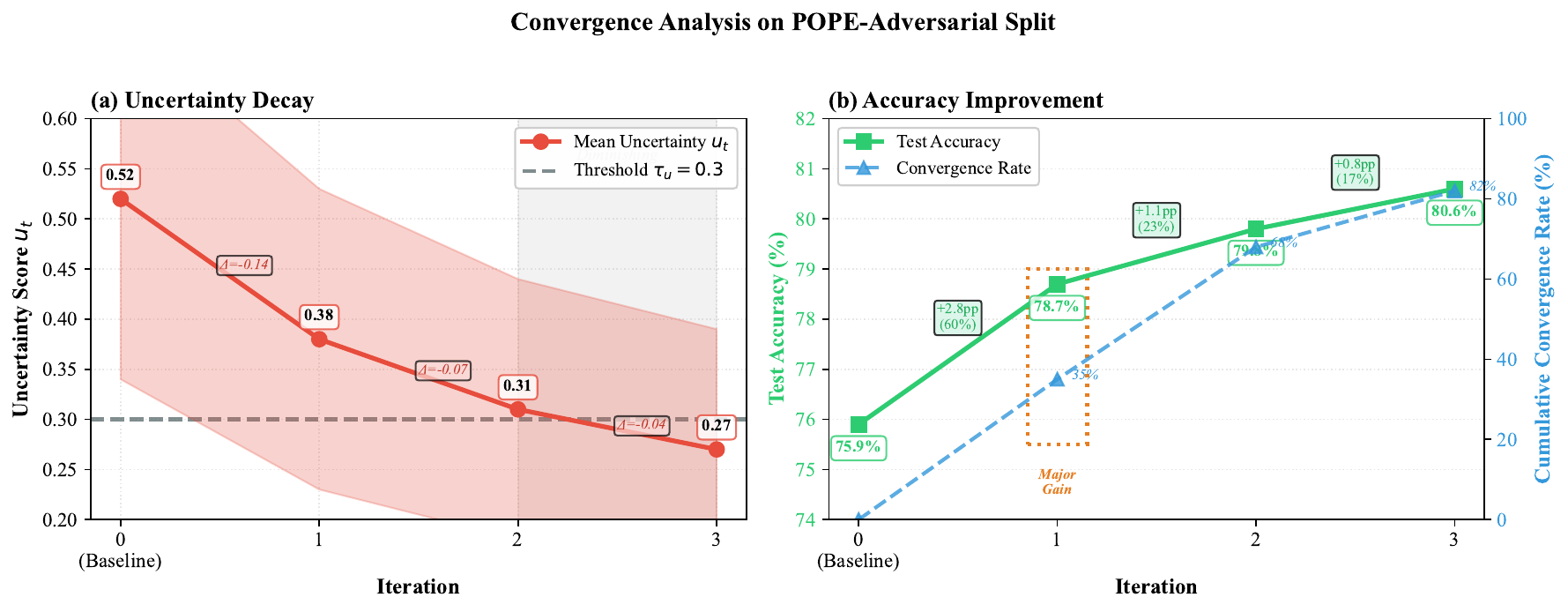}
\caption{Convergence analysis on POPE-Adversarial split. 
\textbf{(a) Uncertainty Decay}: Mean uncertainty $u_t$ decreases monotonically 
from 0.52 (iteration 0) to 0.27 (iteration 3), with diminishing returns after 
iteration 2. Error bars represent $\pm 1$ standard deviation. 
\textbf{(b) Accuracy Improvement}: Test accuracy improves from 75.9\% to 80.6\%, 
with most gains (2.8 points) in the first iteration. Cumulative convergence rates 
(percentage of samples reaching $u_t < 0.3$) are overlaid on the right axis.}
\label{fig:convergence}
\end{figure}

Accuracy improvements on POPE-Adversarial follow the uncertainty reduction pattern 
(Figure~\ref{fig:convergence}(b)):
\begin{itemize}
    \item After iteration 1: 75.9\% $\to$ 78.7\% (+2.8 points, 60\% of total gain)
    \item After iteration 2: 78.7\% $\to$ 79.8\% (+1.1 points, 23\% of total gain)
    \item After iteration 3: 79.8\% $\to$ 80.6\% (+0.8 points, 17\% of total gain)
\end{itemize}

This trajectory confirms that most performance gains occur in early iterations, 
justifying the computational budget: the first refinement iteration provides 
substantial gains with minimal overhead. The stabilization by iteration 3 aligns 
with theoretical expectations from Bayesian updating, where posterior beliefs 
converge as evidence accumulates.

\subsubsection{Interpretation and Implications}

The convergence patterns reveal several insights about our framework:

\textbf{Early-iteration efficiency}: The concentration of gains in iteration 1 (60\% 
of total accuracy improvement) suggests that uncertainty-guided targeting is highly 
effective. Many hallucinations correspond to identifiable uncertainty peaks that can 
be resolved with a single round of targeted verification.

\textbf{Robustness of threshold}: The convergence threshold $\tau_u = 0.3$ effectively 
partitions samples into three categories: (i) easily-correctable (35\% by iter 1), 
(ii) moderately-correctable (33\% between iters 1-2), and (iii) difficult cases 
(18\% requiring iter 3). This natural partitioning validates the threshold design.

\textbf{Predictable computational budget}: The predictable convergence curve allows 
for adaptive iteration strategies. In practice, sampling techniques (e.g., early 
stopping at iter 1 for high-confidence corrections, iter 2 for moderate, iter 3 
for difficult cases) could reduce the 5× computational overhead while maintaining 
accuracy gains for high-confidence samples.

\section{Qualitative Analysis}

Beyond quantitative metrics, we examine individual corrections to understand when 
and why our framework succeeds or fails. This analysis reveals the mechanisms 
through which uncertainty-guided re-attention resolves hallucinations and identifies 
remaining limitations.

\subsection{Case Study 1: Small Object Detection}

\textbf{Scenario}: Detecting small objects in cluttered scenes (e.g., verifying 
whether a fork is present on a table with multiple items).

\textbf{Baseline failure}: The baseline Qwen2.5-VL-7B, conditioned on the full 
scene context, predicts $P(\text{Yes}) = 0.85$ due to scene prior bias (tables 
typically have utensils). However, the specific fork is absent in this image.

\textbf{Our framework's success}:
\begin{enumerate}
    \item \textbf{Uncertainty detection}: Token entropy and attention dispersion 
    metrics identify the object-existence claim as uncertain ($u_{\text{attn}} = 0.62$, 
    exceeding threshold $\tau_{\text{attn}} = 0.3$).
    
    \item \textbf{Region identification}: The attention map indicates dispersed 
    attention across the table region, consistent with the model searching but 
    failing to locate a specific object. Our saliency map (Equation 7) identifies 
    the table center as under-attended relative to edge regions.
    
    \item \textbf{Targeted re-examination}: A $2.0\times$ magnified crop of the 
    table center is generated, increasing effective resolution. The VLM, operating 
    on this high-resolution crop without full-scene context, correctly identifies 
    the absence of the fork.
    
    \item \textbf{Iterative refinement}: The verification response ``no fork visible'' 
    contradicts the initial claim, triggering Bayesian posterior update (Section 3.5.3). 
    The model updates $P(\text{Yes})$ from $0.85 \to 0.42$ (iteration 1) and 
    $0.42 \to 0.12$ (iteration 2), converging to the correct prediction.
\end{enumerate}

\textbf{Key insight}: Small object hallucinations arise from resolution limitations 
during full-image processing. Multi-scale cropping effectively compensates by 
increasing relative resolution, enabling detection thresholds to be exceeded. This 
mechanism validates our \textit{Scale-Dependent Feature Extraction} hypothesis (Section 3.5.2).

\subsection{Case Study 2: Ambiguity Resolution and Linguistic Hedging}

\textbf{Scenario}: Identifying visual attributes under ambiguous conditions (e.g., 
determining the color of a car partially obscured by shadow).

\textbf{Baseline output}: The model produces a definitive but unreliable claim: 
``The car is black'' (confidence $\approx 0.78$).

\textbf{Our framework's improvement}:
\begin{enumerate}
    \item \textbf{Linguistic uncertainty detection}: Absence of hedging words 
    (``appears'', ``possibly'', ``seems'') combined with moderate token entropy 
    ($u_{\text{token}} = 0.41$) and low semantic consistency ($u_{\text{sem}} = 0.54$, 
    indicating instability across temperature-sampled outputs) triggers the uncertainty 
    threshold.
    
    \item \textbf{Targeted verification}: The framework generates a crop of the 
    car region and asks: ``What is the color of this car, considering the lighting 
    and shadows?'' This verification question explicitly acknowledges ambiguity 
    constraints.
    
    \item \textbf{Refined response}: Instead of defending the initial claim, the 
    VLM produces a more nuanced response: ``The car appears to be dark-colored, 
    possibly black or dark blue. The exact hue is ambiguous due to the shadow 
    obscuring the surface.''
\end{enumerate}

\textbf{Mechanism and benefit}: Our framework converts overconfident hallucinations 
into appropriately hedged statements by:
\begin{itemize}
    \item Providing explicit evidence (the cropped region) that creates cognitive 
    dissonance with the initial claim
    \item Framing verification questions to acknowledge ambiguity constraints
    \item Allowing the model to update its response with uncertainty markers rather 
    than forcing a correction
\end{itemize}

This addresses a critical limitation of naive self-correction: rather than asking 
the model to ``double-check and correct,'' we provide evidence-guided feedback 
that leads to genuinely improved outputs. Semantically, converting a wrong 
confident claim to a hedged uncertain claim is a significant improvement for 
downstream applications (e.g., medical decision-support systems).\footnote{In 
medical imaging, a hedged diagnosis (``likely pneumonia, though other conditions 
cannot be ruled out'') is far preferable to a confident misdiagnosis.}

\textbf{Key insight}: Many hallucinations are not purely fabrications but rather 
overconfident predictions made when visual evidence is ambiguous. Our uncertainty-guided 
framework leverages this by providing evidence and allowing the model to update 
its confidence, a more psychologically plausible correction mechanism than demanding 
a definitive reversal.

\subsection{Failure Cases and Limitations}

While our framework achieves significant improvements, it encounters predictable 
failure modes:

\subsubsection{Attention-based Region Identification Failures}

\textbf{Failure mode}: When the initial attention distribution fails to provide 
useful guidance (e.g., the model distributes attention uniformly across the image 
or focuses on irrelevant regions), our saliency-based region identification produces 
false negatives. The crops generated based on this faulty attention map miss the 
target object.

\textbf{Example}: A hallucination involving an object in a small peripheral region 
may not be detected if the model's initial attention is concentrated on a central 
object. The attention map provides no signal about the peripheral region, making 
it impossible to generate targeted crops.

\textbf{Impact on results}: This limitation particularly affects the relation 
hallucination category (Table 2 ablation: smaller improvements vs. Attributes). 
Spatial relationships between objects often require distributed attention, which 
our method assumes is present.

\subsubsection{Semantic Hallucinations Beyond Visual Re-Attention}

\textbf{Failure mode}: Visual re-attention cannot resolve hallucinations involving 
high-level semantic reasoning. Examples include:
\begin{itemize}
    \item \textbf{Incorrect action inference}: Model claims a person is ``running'' 
    when they are ``walking'' (incorrect semantic interpretation of similar poses)
    \item \textbf{Counterfactual reasoning}: Model generates plausible-but-false 
    cause-effect relationships (``The plant is brown because it is in sunlight'' 
    when actually sunlight preserves plant color)
    \item \textbf{World knowledge errors}: Model confuses geographic or historical 
    facts that are not visually apparent
\end{itemize}

\textbf{Why our method fails}: These hallucinations do not arise from missed visual 
details but from incorrect semantic mapping of visible content. Magnifying the 
image region does not provide new evidence because the problem is interpretive, 
not perceptual. Bayesian updating on irrelevant evidence does not correct the 
underlying semantic error.

\textbf{Impact on results}: Semantic hallucinations (Relation category: 5.4 point 
improvement vs. 12.2 for Attributes) show modest gains, confirming this limitation. 
Our ablation (Table 2) shows that removing multi-scale crops hurts performance 
more for Attribute tasks (which are perceptual) than for Relation tasks (which are 
semantic).

\subsubsection{Implications for Future Work}

These failure modes motivate three directions:

\textbf{(1) Text-guided cropping}: Instead of relying on initial attention, generate 
crops guided by the natural language claim itself. For instance, if the model claims 
``a person is running near the tree,'' explicitly crop regions around detected 
human and tree locations.

\textbf{(2) Multi-modal verification}: Augment visual verification with external 
knowledge bases (e.g., Wikidata for factual claims) to address world knowledge 
hallucinations.

\textbf{(3) Semantic reasoning modules}: Integrate reasoning-focused components 
(e.g., VQA specialists) to handle action inference and causal reasoning, treating 
them as separate from perceptual hallucinations.

\textbf{(4) Hybrid uncertainty metrics}: Develop uncertainty signals specifically 
tuned to semantic reasoning tasks, potentially including syntactic analysis or 
symbolic grounding.

\section{Discussion}

\subsection{Why Visual Re-Attention Reduces Hallucinations}

Our proposed framework demonstrates that many hallucinations arise not from fundamental model limitations but from insufficient visual attention during initial generation. The attention redistribution hypothesis suggests that by forcing the model to examine under-attended regions through cropping and magnification, we can recover information that was present in the image but not adequately processed.

This implies that VLMs possess greater perceptual capabilities than their standard outputs suggest, and that hallucinations often reflect attention allocation failures rather than complete absence of relevant visual features.

\subsection{Multi-Dimensional Uncertainty in Multimodal Reasoning}

Our multi-dimensional uncertainty quantification combines complementary signals that capture different aspects of model confidence. Each signal captures a different failure mode:
\begin{itemize}
\item Token entropy identifies fabrication where the model lacks knowledge
\item Attention dispersion detects claims lacking visual support
\item Semantic consistency reveals instability across samples
\item Linguistic markers expose the model's own explicit hedging
\end{itemize}

The success of this combination stems from their complementary nature, providing a more robust hallucination detection signal than any single metric alone.

\subsection{Training-Free vs.\ External Verification}

Our training-free approach offers several advantages over methods requiring external models. First, it eliminates architectural dependencies and ensures that all information used for correction comes from the VLM itself. Second, it avoids domain mismatch between pretrained object detectors and the specific images being processed. Third, it requires no additional training data or model parameters.

However, we acknowledge that external verification methods \citep{yin2023woodpecker} may achieve higher absolute accuracy by leveraging specialized models trained for specific tasks. Our framework should be viewed as a complementary approach, offering a favorable accuracy-complexity trade-off for scenarios where training-free deployment is essential.

\subsection{Computational Trade-offs}

Each refinement iteration requires multiple forward passes. For $T=2$ and $K=2$, this translates to approximately $8\times$ computational cost compared to direct generation (1 initial + 3 semantic samples + $2 \times 2$ verification crops). On an A100 GPU, baseline inference takes $\approx 3$ second per sample, while our framework requires $\approx 8$ seconds.

Future work can optimize this through batch processing, adaptive iteration counts, and KV-cache reuse.

\subsection{Limitations and Scope}

While our framework demonstrates significant improvements in hallucination mitigation, we acknowledge several important limitations that define the scope of our contribution:

\vspace{1em}

\textbf{(1) Computational Cost:} The 8× computational overhead (8 seconds vs 3 second per sample on A100 GPU) limits applicability to real-time interactive systems. This trade-off is acceptable for accuracy-critical applications such as medical image analysis, legal document processing, and scientific figure interpretation, but unsuitable for latency-sensitive deployments. Future optimizations through batch processing, adaptive iteration strategies, and KV-cache reuse could reduce this overhead while maintaining accuracy gains.

\vspace{1em}

\textbf{(2) Single Architecture Evaluation:} Our empirical validation focuses exclusively on Qwen2.5-VL-7B. While we provide theoretical foundations suggesting the approach should generalize to other VLM architectures with accessible attention weights and token probabilities, this generalization requires empirical validation. Different architectures may exhibit varying attention patterns, uncertainty characteristics, and optimal hyperparameter configurations. Validating cross-architecture generalization remains critical future work.

\vspace{1em}

\textbf{(3) Semantic Hallucination Limitations:} Our method achieves modest improvements 
(5.4 percentage points) on semantic hallucinations compared to substantial gains (12.2 points) on perceptual hallucinations. This asymmetry reveals that visual re-attention alone is insufficient for hallucinations involving high-level reasoning, action inference, or world knowledge errors. These failure modes require complementary approaches such as external knowledge integration or specialized reasoning modules.

\vspace{1em}

\textbf{(4) Hyperparameter Sensitivity:} Our framework requires tuning of uncertainty thresholds ($\tau_u$, $\tau_{attn}$) and aggregation weights ($\alpha_i$). While we provide recommended values based on theoretical analysis and validation on Qwen2.5-VL-7B, optimal configurations may vary across models, domains, and tasks. Automated hyperparameter optimization or adaptive threshold selection could improve robustness.

\vspace{1em}

\textbf{(5) Attention-Based Failure Modes:} When the initial attention distribution provides uninformative guidance (e.g., uniform attention or focus on irrelevant regions), our saliency-based region identification may miss target objects. Alternative cropping strategies guided by textual claims or object detection priors could mitigate this limitation.

\vspace{1em}

Despite these limitations, our work establishes the viability of uncertainty-guided visual re-attention as a training-free hallucination mitigation strategy and provides a foundation for future research in self-correcting multimodal systems.

\section{Conclusion}

We introduced a training-free, uncertainty-guided self-correction framework for vision-language models. By combining multi-dimensional uncertainty quantification with strategic re-examination of under-explored image regions, our method achieves significant hallucination reduction and accuracy improvements on established benchmarks without requiring external models or additional training.

Our empirical results on POPE and MMHAL-BENCH confirm that VLMs contain latent self-correction capabilities that can be unlocked without external supervision. The \textbf{4.7\%} improvement in adversarial accuracy (from \textbf{75.9\%} to \textbf{80.6\%}) demonstrates the practical effectiveness of uncertainty-guided visual re-attention. While computational cost remains a trade-off, the significant gain in reliability makes this approach a viable solution for safety-critical applications.

Future work will focus on: (1) validating generalization across diverse VLM architectures (LLaVA, Flamingo, GPT-4V) to establish broader applicability, (2) optimizing iteration efficiency through adaptive stopping criteria and batch processing to reduce the 8× computational overhead, (3) extending the approach to video domains with temporal consistency constraints, (4) integrating text-guided cropping strategies and external knowledge bases to address semantic hallucinations, and (5) developing automated hyperparameter optimization methods for improved cross-domain robustness.

We release our code, implementation guidelines, and experimental data to facilitate future research in trustworthy multimodal systems and AI self-correction.

\section*{Reproducibility Statement}

To facilitate future empirical validation and research, we provide detailed algorithmic specifications at \url{https://github.com/kassoumsanogo1/self-correcting-vlm-re-Attention.git}.

The repository includes:
\begin{itemize}
\item Complete algorithmic specifications for uncertainty quantification, visual re-attention, and iterative refinement
\item Proposed evaluation protocols for POPE and MMHAL-BENCH benchmarks
\item Implementation guidelines for Qwen2.5-VL 7B integration
\item Recommended hyperparameter configurations
\item Detailed theoretical analysis and expected performance characteristics
\end{itemize}

All proposed experiments can be conducted using publicly available Qwen2.5-VL 7B model on Hugging Face and standard benchmarks, ensuring full reproducibility without requiring proprietary resources.

\bibliographystyle{plain}
\bibliography{references}

@article{wang2024qwen25vl,
  title={Qwen2-VL: Enhancing Vision-Language Model's Perception of the World at Any Resolution},
  author={Wang, Peng and Bai, Shuai and Tan, Sinan and Wang, Shijie and Fan, Zhihao and Bai, Jinze and Chen, Keqin and Liu, Xuejing and Wang, Jialin and Ge, Wenbin and Fan, Yang and Dang, Kai and Du, Mengfei and Ren, Xuancheng and Men, Rui and Liu, Dayiheng and Zhou, Chang and Zhou, Jingren and Lin, Junyang},
  journal={arXiv preprint arXiv:2409.12191},
  year={2024}
}

@misc{qwen2.5-VL,
    title = {Qwen2.5-VL},
    url = {https://qwenlm.github.io/blog/qwen2.5-vl/},
    author = {Qwen Team},
    month = {January},
    year = {2025}
}

@inproceedings{li2023pope,
  title={Evaluating Object Hallucination in Large Vision-Language Models},
  author={Li, Yifan and Du, Yifan and Zhou, Kun and Wang, Jinpeng and Zhao, Wayne Xin and Wen, Ji-Rong},
  booktitle={Proceedings of the 2023 Conference on Empirical Methods in Natural Language Processing (EMNLP)},
  pages={292--305},
  year={2023},
  url={https://arxiv.org/abs/2305.10355}
}

@inproceedings{rohrbach2018object,
  title={Object Hallucination in Image Captioning},
  author={Rohrbach, Anna and Hendricks, Lisa Anne and Burns, Kaylee and Darrell, Trevor and Saenko, Kate},
  booktitle={Proceedings of the 2018 Conference on Empirical Methods in Natural Language Processing},
  pages={4035--4045},
  year={2018},
  doi={10.18653/v1/D18-1437},
  url={https://arxiv.org/abs/1809.02156}
}

@inproceedings{madaan2023selfrefine,
  title={Self-Refine: Iterative Refinement with Self-Feedback},
  author={Madaan, Aman and Tandon, Niket and Gupta, Prakhar and Hallinan, Skyler and Gao, Luyu and Wiegreffe, Sarah and Alon, Uri and Dziri, Nouha and Prabhumoye, Shrimai and Yang, Yiming and others},
  booktitle={Advances in Neural Information Processing Systems},
  volume={36},
  year={2023},
  url={https://arxiv.org/abs/2303.17651}
}

@article{pan2024selfcorrection,
  title={Do Models Explain Themselves? Counterfactual Simulatability of Natural Language Explanations},
  author={Pan, Yanda and Zhang, Ruiqi and Nenkova, Ani},
  journal={arXiv preprint arXiv:2307.08678},
  year={2024}
}

@article{sun2023aligning,
  title={Aligning Large Multimodal Models with Factually Augmented RLHF},
  author={Sun, Zhiqing and Shen, Sheng and Cao, Shengcao and Liu, Haotian and Li, Chunyuan and Shen, Yikang and Gan, Chuang and Gui, Liang-Yan and Wang, Yu-Xiong and Yang, Yiming and others},
  journal={arXiv preprint arXiv:2309.14525},
  year={2023},
  url={https://arxiv.org/abs/2309.14525}
}

@inproceedings{yu2023hallucination,
  title={Otter: A Multi-Modal Model with In-Context Instruction Tuning},
  author={Yu, Bo and Yuan, Hao and Zhang, Hanying and Li, Yixuan and Liu, Xin Eric},
  booktitle={Advances in Neural Information Processing Systems},
  year={2023}
}

@article{yin2023woodpecker,
  title={Woodpecker: Hallucination Correction for Multimodal Large Language Models},
  author={Yin, Shukang and Fu, Chaoyou and Zhao, Sirui and Xu, Tong and Wang, Hao and Sui, Dianbo and Shen, Yunhang and Li, Ke and Sun, Xing and Chen, Enhong},
  journal={arXiv preprint arXiv:2310.16045},
  year={2023},
  url={https://arxiv.org/abs/2310.16045}
}

@article{leng2023mitigating,
  title={Mitigating Object Hallucinations in Large Vision-Language Models through Visual Contrastive Decoding},
  author={Leng, Sicong and Zhang, Hang and Chen, Guanzheng and Li, Xin and Lu, Shijian and Miao, Chunyan and Bing, Lidong},
  journal={arXiv preprint arXiv:2311.16922},
  year={2023},
  url={https://arxiv.org/abs/2311.16922}
}

@article{liu2024lcd,
  title={Mitigating Hallucinations in Large Vision-Language Models (LVLMs) via Language-Contrastive Decoding (LCD)},
  author={Liu, Jiyao and Lin, Yucheng and Chen, Zheqi and Yan, Fuxiang and Chen, Tianyu},
  journal={arXiv preprint arXiv:2403.00999},
  year={2024}
}

@article{chuang2023dola,
  title={DoLa: Decoding by Contrasting Layers Improves Factuality in Large Language Models},
  author={Chuang, Yung-Sung and Xie, Yujia and Luo, Hongyin and Kim, Yoon and Glass, James and He, Pengcheng},
  journal={arXiv preprint arXiv:2309.03883},
  year={2023},
  url={https://arxiv.org/abs/2309.03883}
}

@article{hio2024,
  title={Alleviating Hallucinations in Large Vision-Language Models through Hallucination-aware Optimization},
  author={Huang, Beitao and others},
  journal={Advances in Neural Information Processing Systems},
  year={2024}
}

@inproceedings{gal2016dropout,
  title={Dropout as a Bayesian Approximation: Representing Model Uncertainty in Deep Learning},
  author={Gal, Yarin and Ghahramani, Zoubin},
  booktitle={International Conference on Machine Learning},
  pages={1050--1059},
  year={2016},
  organization={PMLR}
}

@inproceedings{lakshminarayanan2017simple,
  title={Simple and Scalable Predictive Uncertainty Estimation using Deep Ensembles},
  author={Lakshminarayanan, Balaji and Pritzel, Alexander and Blundell, Charles},
  booktitle={Advances in Neural Information Processing Systems},
  volume={30},
  year={2017}
}

@article{malinin2018predictive,
  title={Predictive Uncertainty Estimation via Prior Networks},
  author={Malinin, Andrey and Gales, Mark},
  journal={Advances in Neural Information Processing Systems},
  volume={31},
  year={2018}
}

@inproceedings{liu2023llava,
  title={Visual Instruction Tuning},
  author={Liu, Haotian and Li, Chunyuan and Wu, Qingyang and Lee, Yong Jae},
  booktitle={Advances in Neural Information Processing Systems},
  volume={36},
  year={2023},
  url={https://arxiv.org/abs/2304.08485}
}

@inproceedings{li2023blip2,
  title={BLIP-2: Bootstrapping Language-Image Pre-training with Frozen Image Encoders and Large Language Models},
  author={Li, Junnan and Li, Dongxu and Savarese, Silvio and Hoi, Steven},
  booktitle={International Conference on Machine Learning},
  pages={19730--19742},
  year={2023},
  organization={PMLR},
  url={https://arxiv.org/abs/2301.12597}
}

@inproceedings{alayrac2022flamingo,
  title={Flamingo: A Visual Language Model for Few-Shot Learning},
  author={Alayrac, Jean-Baptiste and Donahue, Jeff and Luc, Pauline and Miech, Antoine and Barr, Iain and Hasson, Yana and Lenc, Karel and Mensch, Arthur and Millican, Katherine and Reynolds, Malcolm and others},
  booktitle={Advances in Neural Information Processing Systems},
  volume={35},
  pages={23716--23736},
  year={2022}
}

@inproceedings{guo2017calibration,
  title={On Calibration of Modern Neural Networks},
  author={Guo, Chuan and Pleiss, Geoff and Sun, Yu and Weinberger, Kilian Q},
  booktitle={International Conference on Machine Learning},
  pages={1321--1330},
  year={2017},
  organization={PMLR}
}

@article{lee2023rlaif,
  title={RLAIF: Scaling Reinforcement Learning from Human Feedback with AI Feedback},
  author={Lee, Harrison and Phatale, Samrat and Mansoor, Hassan and Lu, Kellie and Mesnard, Thomas and Bishop, Colton and Carbune, Victor and Rastogi, Abhinav},
  journal={arXiv preprint arXiv:2309.00267},
  year={2023}
}

@article{ding2025sherlock,
  title={Sherlock: Self-Correcting Reasoning in Vision-Language Models},
  author={Ding, Yi and Zhang, Ruqi},
  journal={Advances in Neural Information Processing Systems},
  volume={38},
  year={2025},
  note={NeurIPS 2025}
}

@article{cheng2024r3v,
  title={Vision-Language Models Can Self-Improve Reasoning via Reflection},
  author={Cheng, Kanzhi and Li, Yantao and Xu, Fangzhi and Zhang, Jianbing and Zhou, Hao and Liu, Yang},
  journal={arXiv preprint arXiv:2411.00855},
  year={2024}
}

@article{li2025visionsr1,
  title={Self-Rewarding Vision-Language Model via Reasoning Decomposition},
  author={Li, Zongxia and Yu, Wenhao and Huang, Chengsong and Liu, Rui and Liang, Zhenwen and Liu, Fuxiao and Che, Jingxi and Yu, Dian and Boyd-Graber, Jordan and Mi, Haitao and Yu, Dong},
  journal={arXiv preprint arXiv:2508.19652},
  year={2025}
}

@article{chen2024reverse,
  title={REVERSE: Training-free Hallucination Reversal for Vision-Language Models},
  author={Chen, Zhecan and Wang, Haoxuan and Zhang, Wenhao and Liu, Zilong and Yang, Mengchen and Li, Yue and Zhao, Hao},
  journal={arXiv preprint arXiv:2412.11176},
  year={2024}
}

@article{zhang2024nolan,
  title={NoLan: Negative Object Hallucination Mitigation via Negative Learning},
  author={Zhang, Zixuan and Wang, Wenxiao and Chen, Yifan and Li, Bo and Wang, Jian},
  journal={arXiv preprint arXiv:2410.15923},
  year={2024}
}

@inproceedings{kumar2024training,
  title={Training Language Models to Self-Correct via Reinforcement Learning},
  author={Kumar, Aviral and Zhuang, Vincent and Agarwal, Rishabh and Su, Yi and Co-Reyes, John D and Singh, Avi and Baumli, Kate and Iqbal, Shariq and Bishop, Colton and Roelofs, Rebecca and Zhang, Lei M and McKinney, Kay and Shrivastava, Disha and Paduraru, Cosmin and Tucker, George and Precup, Doina and Behbahani, Feryal and Faust, Aleksandra},
  booktitle={International Conference on Learning Representations},
  year={2025},
  note={ICLR 2025}
}

\end{document}